\documentclass[journal, twoside]{./IEEEtran} 

\usepackage[table]{xcolor}
\definecolor{Light1}{rgb}{0.98, 0.95, 0.90}
\definecolor{Light2}{rgb}{0.98, 0.98, 0.93}
\definecolor{Light3}{rgb}{0.98, 0.98, 1}
\definecolor{Light4}{rgb}{0.93, 0.98, 0.98}

\usepackage{graphics} 
\usepackage{epsfig} 
\usepackage{times} 
\usepackage{mathtools}

\usepackage{booktabs}

\usepackage{subcaption}

\usepackage{tikz}
\usepackage{ifthen}
\usetikzlibrary{decorations.pathreplacing, calc, backgrounds}
\usetikzlibrary{shapes, shadows.blur, arrows.meta, positioning}
\usepackage{pdfrender}
\usepackage{tikzsymbols}
\usepackage{pgfplots}
\usepackage{pgfplotstable}
\pgfplotsset{compat=1.11}

\makeatletter
\newcommand{\gettikzxy}[3]{%
  \tikz@scan@one@point\pgfutil@firstofone#1\relax
  \edef#2{\the\pgf@x}%
  \edef#3{\the\pgf@y}%
}
\makeatother

\usepackage{multirow}
\usepackage{makecell}

\tikzstyle{generalStyle}=[blur shadow={shadow blur steps=7}]

\newcommand{\drawIndividualFrustum}[7]{
	\set{\xTempFilledBlock}{#1};
	\set{\yTempFilledBlock}{#2};
	\set{\zTempFilledBlock}{-#3};
	
	\set{\leftHeightTempFilledBlock}{#4};
	\set{\rightHeightTempFilledBlock}{#5};
	\set{\widthTempFilledBlock}{#6};

	\pgfmathparse{\leftHeightTempFilledBlock>\rightHeightTempFilledBlock};
	\set{\isLeftBigger}{\pgfmathresult};
	\ifthenelse{\isLeftBigger=1}{
		\set{\leftAdd}{0};
		\set{\rightAdd}{(\leftHeightTempFilledBlock - \rightHeightTempFilledBlock) * 0.5};
	}{
		\set{\leftAdd}{(\rightHeightTempFilledBlock - \leftHeightTempFilledBlock) * 0.5};
		\set{\rightAdd}{0};
	};
	
	\draw [#7] (\xTempFilledBlock, \yTempFilledBlock+\leftAdd, \zTempFilledBlock) -- 
	(\xTempFilledBlock+\widthTempFilledBlock, \yTempFilledBlock + \rightAdd, \zTempFilledBlock) -- 
	(\xTempFilledBlock+\widthTempFilledBlock, \yTempFilledBlock+\rightHeightTempFilledBlock+\rightAdd, \zTempFilledBlock) -- 
	(\xTempFilledBlock, \yTempFilledBlock+\leftHeightTempFilledBlock+\leftAdd, \zTempFilledBlock) -- cycle;	
}

\newcommand{\drawLegendEntry}[6]{
	\pgfmathsetmacro{\posX}{#1};
	\pgfmathsetmacro{\posY}{#2};
	\pgfmathsetmacro{\width}{#3};
	\pgfmathsetmacro{\useit}{#6};
	\pgfmathsetmacro{\posXX}{\posX + \width};
	\node at (\posX,\posY) (start) {};
	\node[anchor=west] at (\posXX,\posY) (end) {#4}; 
	
	\node at (\posX+0.2,\posY-0.17) (startrect) {};
	\node at (\posXX,\posY+0.17) (endrect)  {};
	\begin{scope}
		\ifthenelse{\useit = 1}{
			\draw[#5, line width=3.5mm, draw opacity=0.25] (start) -- (end);
			
			\draw[#5, line width=0.7mm] (start) -- (end);
		}{
			\draw[#5, line width=0.8mm] (start) -- (end);
		};
	\end{scope}
	
}

\tikzstyle{imgStyle}=[blur shadow={shadow blur steps=7, shadow xshift=1pt, shadow yshift=-1pt}]
\tikzset{
	boxFrame/.style={ minimum width=1.5 cm,%
		minimum height=1.5 cm,%
		align=center}
}

\usepackage{xargs} 
\usepackage{caption}
\usepackage{algorithm} 
\usepackage[noend]{algpseudocode}
\usepackage{bm} 
\usepackage{mathbbol}

\usepackage{varwidth}
\usepackage{subcaption}

\newcommand{\set}[2]{\pgfmathsetmacro{#1}{#2}}

\makeatletter
\DeclareRobustCommand\onedot{\futurelet\@let@token\@onedot}
\def\@onedot{\ifx\@let@token.\else.\null\fi\xspace}

\def\eg{\emph{e.g}\onedot}

\def\etal{\emph{et al}\onedot}
\makeatother

\DeclareMathOperator*{\argmin}{argmin}

\usepackage[breaklinks,colorlinks]{hyperref}

\usepackage[capitalize]{cleveref}
\Crefname{section}{Sec.}{Secs.}
\Crefname{section}{Section}{Sections}
\Crefname{table}{Table}{Tables}
\Crefname{table}{Tab.}{Tabs.}

\usepackage{threeparttable}

\algnewcommand{\LeftComment}[1]{\Statex \(\triangleright\) #1}

\usepackage{cite}
\usepackage{amsmath,amssymb,amsfonts}
\def\BibTeX{{\rm B\kern-.05em{\sc i\kern-.025em b}\kern-.08em
    T\kern-.1667em\lower.7ex\hbox{E}\kern-.125emX}}

\Crefname{equation}{Eq.}{Eqs.} 
\Crefname{figure}{Fig.}{Figs.}

\graphicspath{{images-src/}{images-bin/}} 

\definecolor{dlrprim1}{HTML}{000000} 
\definecolor{dlrprim2}{HTML}{666666} 
\definecolor{dlrprim3}{HTML}{b9cad2}
\definecolor{dlrprim4}{HTML}{ffffff} 

\definecolor{dlrblue1}{HTML}{00658b} 
\definecolor{dlrblue2}{HTML}{3b98cb}
\definecolor{dlrblue3}{HTML}{6cb9dc}
\definecolor{dlrblue4}{HTML}{a7d3ec}
\definecolor{dlrblue5}{HTML}{d1e8fa}

\definecolor{dlryellow1}{HTML}{d2ae3d}  
\definecolor{dlryellow2}{HTML}{f2cd51} 
\definecolor{dlryellow3}{HTML}{f8de53}
\definecolor{dlryellow3}{HTML}{fcea7a}
\definecolor{dlryellow3}{HTML}{fff8be}

\definecolor{dlrgreen1}{HTML}{82a043} 
\definecolor{dlrgreen2}{HTML}{a6bf51}
\definecolor{dlrgreen3}{HTML}{cad55c}
\definecolor{dlrgreen4}{HTML}{d9df78}
\definecolor{dlrgreen5}{HTML}{e6eaaf}

\definecolor{dlrgray1}{HTML}{666666} 
\definecolor{dlrgray2}{HTML}{868585}
\definecolor{dlrgray3}{HTML}{b1b1b1}
\definecolor{dlrgray4}{HTML}{cfcfcf}
\definecolor{dlrgray5}{HTML}{ebebeb}

\usepackage[textsize=scriptsize]{todonotes} 
\renewcommand{\todo}[1]{}

\tikzset{
	cross/.pic = {
		\draw[rotate = 45, line width=0.75pt] (-#1,0) -- (#1,0);
		\draw[rotate = 45, line width=0.75pt] (0,-#1) -- (0, #1);
	}
}

\newcommand{\greencheckmark}{\tikz{\path (-0.1, 0.) -- (-0.1, 0); \fill[scale=0.4, color=green](0,.35) -- (.25,0) -- (1,.7) -- (.25,.15) -- cycle;}}
\newcommand{\yellowcheckmark}{\tikz{\path (-0.1, 0.) -- (-0.1, 0); \fill[scale=0.4] (0,.35) -- (.25,0) -- (1,.7) -- (.25,.15) -- cycle;}}
\newcommand{\xSymbol}{\tikz\path (0, 0) pic[red] {cross=4pt};}
\newcommand{\yellowxSymbol}{\tikz\path (0, 0) pic {cross=4pt};}

\usepackage{orcidlink}

\newcommand{\rmcaffiliation}{German Aerospace Center (DLR),
	Institut of Robotics and Mechatronics (RMC),
	M\"unchener Str. 20, 82234 We\ss ling, Germany. }

\IEEEoverridecommandlockouts                             

\newcommand{\bmu}{\bm{\mu}}

\newcommand{\bk}{\bm{k}}
\newcommand{\bK}{\bm{K}}
\newcommand{\bxi}{\bm{\xi}}
\newcommand{\bs}{\bm{s}}

\newcommand{\bSigma}{\bm{\Sigma}}

\newcommand{\bA}{\bm{A}}
\newcommand{\bb}{\bm{b}}

\newcommand{\timeStep}{t}
\newcommand{\allTimeSteps}{T}
\newcommand{\mean}{\bm{\mu}}
\newcommand{\covariance}{\bm{\Sigma}}

\newcommand{\inputVariable}{\bm{s}}
\newcommand{\outputVariable}{\bm{\xi}}

\newcommand{\threshold}{\gamma}
\newcommand{\thresholdTrajectory}{\threshold_\xi}

\newcommand{\inputDimension}{\mathcal{I}}
\newcommand{\outputDimension}{\mathcal{O}}

\newcommand{\local}{(p)}
\newcommand{\localMean}{\mean^{\local}}
\newcommand{\localCovariance}{\covariance^{\local}}
\newcommand{\amountOfFrames}{P}
\newcommand{\frameIndex}{p}

\newcommand{\frameOrigin}{\bm{b}}
\newcommand{\localFrameOrigin}{\frameOrigin^{\local}}

\newcommand{\frameRotationMatrix}{\bm{A}}
\newcommand{\localFrameRotationMatrix}{\frameRotationMatrix^{\local}}
\newcommand{\frameRotationMatrixAtTimestep}{\bm{A}_{\timeStep}}

\newcommand{\amountOfDatapoints}{M}

\newcommand{\amountOfKMP}{N}
\newcommand{\kmpIndex}{n}

\newcommand{\viapoint}{\bar{v}}
\newcommand{\globalViapoint}{\viapoint}
\newcommand{\localViapoint}{\viapoint^{\local}}
\newcommand{\viapointMean}{\bar{\mean}}
\newcommand{\viapointCovariance}{\bar{\covariance}}
\newcommand{\localViapointMean}{\viapointMean^{\local}}
\newcommand{\localViapointCovariance}{\viapointCovariance^{\local}}
\newcommand{\viapointInput}{\bar{\inputVariable}}

\newcommand{\kmp}{\bm{D}}
\newcommand{\kmpInput}{\inputVariable_{\kmpIndex}}
\newcommand{\localkmpInput}{\kmpInput^{\local}}
\newcommand{\kmpMean}{\mean_{\kmpIndex}}
\newcommand{\localkmpMean}{\kmpMean^{\local}}
\newcommand{\kmpCovariance}{\covariance_{\kmpIndex}}
\newcommand{\localkmpCovariance}{\kmpCovariance^{\local}}
\newcommand{\localkmp}{\Theta^{\local}}
\newcommand{\localkmpVanilla}{\kmp^{\local}}
\newcommand{\tpkmp}{\Theta}

\newcommand{\force}{\bm{F}}
\newcommand{\thresholdForce}{\threshold_{F}} 

\newcommand{\orcidMarkus}{0000-0001-8229-9410}
\newcommand{\orcidAlin}{0000-0001-5343-9074}
\newcommand{\orcidFreek}{0000-0001-9555-9517}
\newcommand{\orcidJoao}{0000-0003-1428-8933}
\newcommand{\addorcidMarkus}{\orcidlink{\orcidMarkus}}
\newcommand{\addorcidAlin}{\orcidlink{\orcidAlin}}
\newcommand{\addorcidFreek}{\orcidlink{\orcidFreek}}
\newcommand{\addorcidJoao}{\orcidlink{\orcidJoao}}

\title{
	Interactive incremental learning of generalizable skills with local trajectory modulation
}

\author{
	Markus Knauer\addorcidMarkus$^{1, 2}$, Alin Albu-Sch\"affer\addorcidAlin$^{1, 2}$, Freek Stulp\addorcidFreek$^{1}$ and Jo\~ao Silv\'erio\addorcidJoao$^{1}$ 
	\thanks{This paper has been accepted at IEEE Robotics and Automation Letters (RA-L).
	This work was partially funded by the DLR project “Factory of the Future Extended (FoF-X)” and the European Union’s Horizon Research and Innovation Program under Grant 101136067 (INVERSE).}
	\thanks{$^{1}$ All authors are with the \rmcaffiliation \texttt{\{first\}.\{last\}@dlr.de}}
	\thanks{$^{2}$ Markus Knauer and Alin Albu-Sch\"affer are with the School of Computation, Information and Technology (CIT), Technical University of Munich (TUM),	Arcisstr. 21, 80333 Munich, Germany. \texttt{m.knauer@tum.de}}
	\thanks{Digital Object Identifier (DOI): \href{https://doi.org/10.1109/LRA.2025.3542209}{10.1109/LRA.2025.3542209}} 
	
}

\begin{document}

	\maketitle

	\begin{abstract}
		The problem of generalization in learning from demonstration (LfD) has received considerable attention over the years, particularly within the context of movement primitives, where a number of approaches have emerged. Recently, two important approaches have gained recognition.
While one leverages via-points to adapt skills locally by modulating demonstrated trajectories, another relies on so-called \textit{task-parameterized} (TP) models that encode movements with respect to different coordinate systems, using a product of probabilities for generalization.
While the former are well-suited to precise, local modulations, the latter aim at generalizing over large regions of the workspace and often involve multiple objects.
Addressing the quality of generalization by leveraging both approaches simultaneously has received little attention.
In this work, we propose an interactive imitation learning framework that simultaneously leverages local and global modulations of trajectory distributions.
Building on the kernelized movement primitives (KMP) framework, we introduce novel mechanisms for skill modulation from direct human corrective feedback.
Our approach particularly exploits the concept of via-points to incrementally and interactively 1) improve the model accuracy locally, 2) add new objects to the task during execution and 3) extend the skill into regions where demonstrations were not provided.
We evaluate our method on a bearing ring-loading task using a torque-controlled, 7-DoF, DLR SARA robot \cite{Iskandar20}.
	\end{abstract}
	
	
	\section{Introduction}
\begin{figure}
	\centering
	\includegraphics[width=0.45\textwidth]{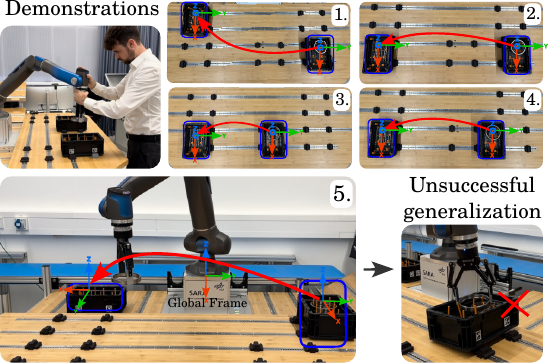}
	\caption{By demonstrating on the same table (top row), users inadvertently introduce low variance along the vertical axis in multiple demonstrations of a ring-loading task, which prevents TP-GMM \cite{Calinon16} from generalizing to different table heights (bottom row).}
	\label{fig_overview_generalization}
\end{figure}

\begin{figure}
	\centering
	\includegraphics[width=0.444\textwidth]{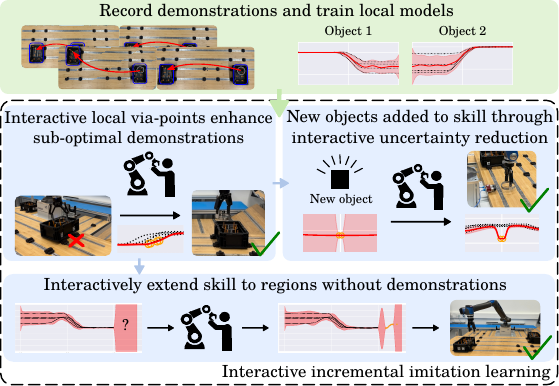}
	\caption{Approach overview. Users provide demonstrations fitted locally to different objects using KMPs (top). The resulting model, TP-KMP, is interactively adjusted to improve generalization and augment the demonstrated skill. An interactive strategy to add via-points locally, with respect to different objects, is the cornerstone of our approach, enabling correction of skills (center-left), addition of new objects (center-right), and skill augmentation (bottom).}
	\label{fig_approach_overview}
\end{figure}

	\IEEEPARstart{T}{ask-parameterized} Gaussian mixture models (TP-GMMs) \cite{Calinon16} are a popular approach to encoding the variations and correlations across multiple demonstrations, facilitating skill generalization. Unlike earlier attempts such as dynamic movement primitives (DMPs) \cite{Ijspeert13} and other probabilistic models for movement primitives like Gaussian mixture models (GMMs) \cite{Calinon07}, TP-GMMs are better suited for adapting to new situations, including those involving multiple objects \cite{Ravichandar20}.
TP-GMMs build local representations of demonstrated trajectories with respect to objects of interest, represented by their poses.
They then generalize the demonstrations to new situations, by formulating generalization as a fusion problem, where each object's local model is weighed against the others in a continuous fashion through a product of Gaussians, generating a trajectory distribution for the robot to track.
Despite their adaptability, TP-GMMs are prone to modeling errors, particularly when imperfect demonstrations introduce ambiguity between objects, which affects generalization when task conditions change, see \cref{fig_overview_generalization}.
In addition, introducing new objects into a learned skill (by defining new task parameters), requires providing new demonstrations and training a new model, when often simple modulations of the original model would suffice.
In this paper, we propose to alleviate these issues via interactive imitation learning \cite{Celemin22}.
We do this by modulating local models from physical user feedback, both improving the generalization of taught skills -- by locally correcting errors -- and permitting their incremental re-use.
Notably, adaptations are applied only locally, with respect to objects of interest, yielding an updated skill model which accurately generalizes new behaviors to new situations.
We argue that in order to modulate skills locally, we require an underlying skill representation that 1) encodes trajectory distributions with both aleatoric and epistemic uncertainties—randomness in data and model knowledge gaps \cite{Huellermeier21}, and 2) allows for trajectory modulation trivially when modifications of the original demonstrations are required.
To achieve this we build on the kernelized movement primitives (KMP) framework \cite{Huang19} (see \cref{sec:background} for a background review).
Although task-parameterization of KMPs is briefly introduced in \cite{Huang19}, the adaptation of local models has received little attention.
We address the adaptation of local models by investigating \textit{when}, \textit{where} and \textit{how} to add via-points.
The result is an interactive learning framework of task-parameterized skills with local trajectory modulation (\cref{sec:approach}) that 1) permits the interactive correction of model imperfections locally, such that corrections `move' with objects, 
2) allows the definition of new task parameters interactively, when tasks change to require new/different objects and 3) ensures compliant interaction when extending skills beyond their initial duration by regulating stiffness based on the epistemic uncertainty, while simultaneously adding via-points.
Figure \ref{fig_approach_overview} provides an overview of our approach.
Our experimental evaluation (\cref{sec:evaluation}) shows that our framework permits users to incrementally build on an initial model of a skill by interactively correcting errors and adding new behaviors in any task phase. We discuss the results and provide conclusions in Sec. \ref{sec:discussion}--\ref{sec:conclusion}.
An overview of the notations and acronyms used is provided in the Additional Material \ref{sec:notations}--\ref{sec:acronyms}.

	\section{Related work}
	\label{sec:relatedWork}
	\subsection{Task-parameterized GMMs and variations}
Several works have emerged both to address limitations of \cite{Calinon16} and introduce new features. In \cite{Huang18} Huang \etal propose to associate confidence factors to the different task parameters, allowing to regulate their influence during task execution.
Zhu \etal \cite{Zhu22} introduce an algorithm to generate new data artificially by sampling from underlying models and re-training the models to improve generalization.
Building on \cite{Huang18}, Sun \etal \cite{Sun23} and Sena \etal \cite{Sena19} introduce strategies to optimize frame relevance given task objectives. Similar re-optimization is required in incremental approaches for TP models, such as \cite{Hoyos16}, which also relies on an underlying GMM.
Recently, Yao \etal \cite{Yao24}, propose to replace the GMM representation by a ProMP \cite{Paraschos13}, introducing the idea of improving generalization by modulating TP models with via-points.
However, via-points are defined globally and projected locally on all frames.
This creates ambiguity when task parameters change, since all frames have high confidence on the via-point positions, requiring the re-definition of via-points every time a task changes.
Although these approaches improve generalization in scenarios with non-moving objects, they do not permit adding new task parameters to existing models and, most importantly, they do not provide interactive capabilities to incrementally refine skills.

\subsection{Interactive imitation learning}
The emergence of interactive imitation learning \cite{Celemin22} highlights the importance of complementing the strengths of classical machine learning methods with interactive mechanisms when acquiring skills in the real world.
Along these lines, Franzese \etal \cite{Franzese2021} introduce the concept of interactive task-frame selection, focusing on object-centered skills in a similar spirit to TP approaches, despite not contemplating the learning of trajectory distributions. Interactive incremental learning approaches in LfD include state-based representations \cite{Saveriano18, Khoramshahi20, jauhri21, Franzese21, Jadav24}, where skills are learned as time-independent autonomous systems and modulated directly in task space by introducing new state-action pairs.
Huang~\etal \cite{Huang19} introduced KMPs, aiming for a probabilistic representation that can be adapted to pass through new via-points after the initial demonstration phase.
Despite introducing a TP formulation of KMP, the authors do not explore the combination of interactive via-point-based modulation with object-centered representations.

\subsection{Our approach and contribution} We extend on the combination of \cite{Calinon16} with KMP-based local trajectory representations, referred to as TP-KMP, leveraging the intuitive addition of via-points after initial demonstration by specifying their location and covariance, as proposed in \cite{Huang19}.
In previous work \cite{Yao24}, via-points were added globally for obstacle avoidance, necessitating re-demonstration when objects were moved.
In contrast, we add via-points locally, allowing them to move with the objects. 
Defining via-points with low covariance increases the importance of the corresponding objects in local models, enhancing the relevance of these via-points in new scenarios.
Our framework uses this property, incorporating human feedback - often investigated for trajectory modulation \cite{Losey18} - to specify via-point locations.
Building on this insight, we further develop an interactive learning framework for defining new task parameters and improving skills incrementally, by leveraging the capabilities of KMPs to encode both aleatoric and epistemic uncertainties \cite{Huellermeier21, Silverio19}. Uncertainty quantifications, important in robotics \cite{fox99, medina15, calandra15}, allow the definition of robot behaviors through uncertainty-aware stiffness regulation strategies \cite{calinon14, medina12}.
In comparison to other approaches like \cite{Fanger16}, we rely on both epistemic and aleatoric uncertainties in a principled manner (\cref{subsec:uncertainties}).

In the domain of incremental learning, approaches like \cite{Zhang2019, spencer2020}, despite allowing for corrections, are either not task-parameterized or require large amounts of training data, making them less intuitive for novice users.
ILoSA \cite{Franzese21} or its extended version \cite{Meszaros22} provide a way to apply partial correction demonstrations locally.
However, due to the missing aleatoric uncertainty a combination of local models is only possible using a heuristics instead of an uncertainty-based fusion like the one used in TP-KMP. In addition, using via-points instead of correction demonstrations is more efficient since the correction only has to be given once.

Finally, compared to DMPs \cite{Ijspeert13}, which offer spatial and temporal invariance for movement scaling, 
our approach additionally provides a greater flexibility by a task-parameterized formulation \cite{Calinon16}. This enables continuous adaptation to multiple objects and purposeful interactions throughout the task, surpassing the fixed goal attractor limitations of DMPs.

\begin{table}[]
	\begin{center}
		\caption{Comparison of related task-parameterized approaches. Our approach models both uncertainties and offers multiple possibilities to adapt motion primitives.}
		\label{tab:comparison_approaches}
			\begin{threeparttable}
				\resizebox{\columnwidth}{!}{%
				\begin{tabular}{l|c|c|c|c|c|c}
					& 
					\multirow{2}{*}{\shortstack{Aleatoric\\uncertainty}} & \multirow{2}{*}{\shortstack{Epistemic\\ uncertainty}} & \multirow{2}{*}{\shortstack{Global\\via-point}}& \multirow{2}{*}{\shortstack{Local\\via-point}} & \multirow{2}{*}{\shortstack{New\\frames}} & \multirow{2}{*}{\shortstack{Interactive\\modulation}} \\ [2.5ex]
					\hline\hline
					TP-GMM/GMR \cite{Calinon16} & \checkmark & - & - & - & -& -\\ 
					TP-ProMP \cite{Yao24} & \checkmark & - & \checkmark & - & - & -\\
					vanilla TP-KMP \cite{Huang19} & \checkmark* & \checkmark* & \checkmark & - & - & -\\
					TP-ILoSA \cite{Meszaros22} & - & \checkmark & \checkmark** & \checkmark** & - & \checkmark \\
					\rowcolor{Light4}
					Our approach & \checkmark & \checkmark & \checkmark & \checkmark & \checkmark & \checkmark \\
				\end{tabular}
				}
				\begin{tablenotes}\footnotesize
					\item[*] available jointly, not used for interaction
					\item[**] Via-points specified simply as new datapoints through corrections
				\end{tablenotes}
			\end{threeparttable}
	\end{center}

\end{table}

In \cref{tab:comparison_approaches} we compare our approach to other task-parameterized approaches on feature level.
As shown in \cite{Ravichandar20} task-parameterized approaches are more suitable for adapting to new situations than non-task-parameterized approaches like \cite{Zhang2019} or \cite{spencer2020}, including those involving multiple objects.
To the best of our knowledge, we present the first task-parameterized approach allowing skill extensions with new frames and the application of local via-points, making it possible to move corrections with their corresponding objects and allowing interactive modulations, where corrections/extensions can be provided and incorporated into the policy online during execution.

	\section{Preliminaries}
	\label{sec:background}
	Let us denote a set of demonstrations by $\{\{\inputVariable_{h,m},\outputVariable_{h,m}\}_{h=1}^H\}_{m=1}^M$, where $\inputVariable\in\mathbb{R}^{\inputDimension}$ and $\outputVariable \in \mathbb{R}^{\outputDimension}$ represent input and output, and $\inputDimension$, $\outputDimension$, $\amountOfDatapoints$, $H$, are the dimensions of input and output, number of demonstrations and trajectory length, respectively.
Similarly to many popular LfD approaches \cite{Calinon07, Huang19, Paraschos13}, we focus on extracting the relationship between $\inputVariable$ and $\outputVariable$ from demonstrations. Depending on the task, $\inputVariable$ is often time or robot state while $\outputVariable$ represents desired poses or velocities. In \cref{sec:background} and \cref{sec:approach} we keep these generic except when concrete examples help explain new concepts.

\subsection{Task parameterized movement models}\label{sec:TP}
In TP models \cite{Calinon16} a \textit{frame} $p=1,\ldots,\amountOfFrames$ is described by so-called \textit{task parameters} $\frameOrigin^{\local}, \frameRotationMatrix^{\local}$, which represent the position and orientation of an object with respect to a common reference frame (e.g. the robot base)\footnote{Task parameters can represent a number of affine transformations commonly found in robotics \cite{Calinon16}.
Here, as in other works \cite{Huang18, Zhu22, Sun23, Yao24}, we use them to represent the coordinate systems of objects of interest.},
where demonstrations are recorded.
Demonstrated outputs are projected locally, into the coordinate systems of different objects, through $\outputVariable^{\local} = \frameRotationMatrix^{\local^{-1}}\left(\outputVariable - \frameOrigin^{\local}\right)$.
Local datasets are modeled probabilistically yielding a Gaussian distribution $\mathcal{N}(\localMean, \localCovariance)$ for every input $\inputVariable$.
During skill execution, local distributions are mapped to the common frame, as both task parameters and inputs change with time $t$, through $\hat{\mean}^{\local}_{\timeStep} = \frameRotationMatrixAtTimestep^{\local} \localMean_{\timeStep} + \localFrameOrigin_{\timeStep}$, $\hat{\covariance}^{\local}_{\timeStep}= \localFrameRotationMatrix_{\timeStep} \localCovariance_{\timeStep} \frameRotationMatrixAtTimestep^{\local^\top}$.
Values of $\frameOrigin^{\local}, \frameRotationMatrix^{\local}$ may differ from those seen during the demonstrations, and, through $\hat{\mean}^{\local}_{\timeStep}$, $\hat{\covariance}^{\local}_{\timeStep}$, each frame provides a model of $\outputVariable$ from its perspective.
The distributions from different coordinate systems are fused by a product of Gaussians, resulting in a new distribution $\mathcal{N}(\mean_{\timeStep}, \covariance_{\timeStep})$, in the common frame, with parameters
\begin{equation}
	\mean_{\timeStep} = \covariance_{\timeStep} \sum^\amountOfFrames_{\frameIndex=1} \hat{\covariance}^{\local^{-1}}_{\timeStep} \hat{\mean}^{\local}_{\timeStep}, \>\> \covariance_{\timeStep} = \left(\sum^\amountOfFrames_{\frameIndex=1} \hat{\covariance}^{\local^{-1}}_{\timeStep}\right)^{-1}.
	\label{eq:gaussian_product}
\end{equation}
The solution of \eqref{eq:gaussian_product} favors models with low variance, being an efficient way to extract features from local models that were consistent across demonstrations, facilitating generalization to new situations.
	\subsection{Kernelized movement primitives (KMPs)}\label{sec:KMP}
KMPs \cite{Huang19} are used in LfD to predict the distribution of $\outputVariable$ given observations of $\inputVariable$.
A KMP is initialized with a \textit{reference trajectory distribution} comprised of $\amountOfKMP$ Gaussians with parameters $\left\lbrace \kmpMean, \kmpCovariance \right\rbrace^\amountOfKMP_{\kmpIndex=1}$, computed from human demonstrations for inputs $\inputVariable_{\kmpIndex=1,\dots,\amountOfKMP}$ using a GMM.
For a test input $\inputVariable^*$, the expectation and covariance of $\outputVariable(\inputVariable^*)$ are given by
\begin{align}
	\mathbb{E}\left[\outputVariable(\inputVariable^*)\right] \! &= \! \bk^*\left(\bK + \lambda_1\covariance\right)^{-1}\mean,\label{eq:kmp_mean}\\
	\mathrm{cov}\left[\outputVariable(\inputVariable^*)\right] \! &= \!\alpha\left(\bk^{**}-\bk^{*}\left(\bK + \lambda_2\bSigma\right)^{-1}\bk^{*\top}\right),\label{eq:kmp_cov}
\end{align}
where 
$\bK = [\hat{\bk}(\inputVariable_1)^\top, \dots, \hat{\bk}(\inputVariable_N)^\top]$,
$\bk^*= \hat{\bk}(\inputVariable^*)$, with
${\hat{\bk}(\inputVariable_i) = \left[\bk(\inputVariable_i,\inputVariable_1),\ldots,\bk(\inputVariable_i,\inputVariable_N)\right]}$, 
${\bk^{**}=\bk(\inputVariable^*,\inputVariable^*)}$, ${\bk(\inputVariable_i,\inputVariable_j)=k(\inputVariable_i,\inputVariable_j)\bm{I}}$, where $\bm{I}$ is an identity matrix, and $k(\inputVariable_i,\inputVariable_j)$ is a kernel function. 
Moreover, $\mean =  \left[\mean_1^\top \ldots \mean_\amountOfKMP^\top\right] \nonumber^\top$, ${\covariance = \mathrm{blockdiag}\left(\covariance_1,\ldots,\covariance_\amountOfKMP\right)}$ and $\lambda_1$, $\lambda_2$, $\alpha$ are hyperparameters.
The kernel matrices are denoted as $\bm{K}, \bm{k}^*$ and $\bm{k}^{**}$.
From \eqref{eq:kmp_mean}--\eqref{eq:kmp_cov} it follows that if, for a certain $\kmpMean$, the covariance $\kmpCovariance$ is small, the expectation at $\kmpInput$ will be close to $\kmpMean$.
This provides a principled way for trajectory modulation.
Indeed, if, for a new input $\viapointInput$, one wants to ensure that the expectation passes through a desired $\viapointMean$, it suffices to manually add the pair $\{\viapointMean,\viapointCovariance\}$ to the reference distribution provided that $\viapointCovariance$ is small enough. This both makes \eqref{eq:kmp_mean} closely match $\viapointMean$ and lowers the covariance \eqref{eq:kmp_cov} to match $\viapointCovariance$.

	\section{Interactive local trajectory modulation with TP-KMP}
	\label{sec:approach}
	Similarly to \cite{Huang19}, we define a local KMP as a model ${\localkmp = \{ \localFrameOrigin,\localFrameRotationMatrix, \localkmpVanilla \}}$  with associated task parameters $ \localFrameOrigin,\localFrameRotationMatrix $ and $\localkmpVanilla = \{\localkmpInput, \localkmpMean, \localkmpCovariance\}^\amountOfKMP_{\kmpIndex=1}$. 
$\bmu^{(p)}$, $\bSigma^{(p)}$ are computed from output data projected locally $\outputVariable^{\local}$, using a GMM.
A TP-KMP is a set of $\amountOfFrames$ local KMPs: $\tpkmp = \{\localkmp\}^\amountOfFrames_{\frameIndex=1}$, where each local KMP generates a distribution $\mathcal{N}(\mean^{\local}, \covariance^{\local})$, computed from \eqref{eq:kmp_mean}--\eqref{eq:kmp_cov}, which is used in \eqref{eq:gaussian_product}.
We introduce an approach to interactively add via-points to local KMPs at any moment of a task, allowing users to intuitively improve models trained on sub-optimal demonstrations.

\subsection{Interactive trajectory modulation with local via-points} \label{subsec:interactive_vp}
In our approach, users add via-points via physical corrections locally, in different object frames, as opposed to in a common global frame.
This enables the adaptation of robot behavior without retraining the model from scratch.
Adding via-points in the relevant local frames has the advantage that corrections `move' with the objects when task conditions change, facilitating generalization.
The via-point mechanism of KMP entails the definition of a small covariance matrix, automatically assigning high importance in the Gaussian product \eqref{eq:gaussian_product} to local KMPs that receive new via-points.

\begin{algorithm}[th]
	\caption{\emph{Trajectory modulation with local via-points}}
	\begin{algorithmic}[1]
		\State{\textbf{Define} external force threshold $\thresholdForce$, distance threshold $\threshold_{\xi}$ and via-point variance $\threshold_{\Sigma}$}
		\State{\textbf{Input}: $\amountOfFrames$ local KMPs $\localkmp = \{\localFrameOrigin, \localFrameRotationMatrix, \localkmpVanilla \}^\amountOfFrames_{\frameIndex=1}$ trained from $\{\{\inputVariable_{h,m},\outputVariable_{h,m}\}_{h=1}^H\}_{m=1}^M$}
		\For{time-step $\timeStep$ in all time-steps $\allTimeSteps$}
			\If{interaction is triggered (see \ref{subsec:interactive_vp})} 
				\LeftComment{Create global via-point}
				\State{$\viapoint_\timeStep = \{\viapointInput = \timeStep, \viapointMean = \outputVariable_{\timeStep}, \viapointCovariance = \threshold_{\Sigma}\bm{I}\}$} 
				\LeftComment{Find $\frameIndex^*$ (see \ref{subsec:interactive_vp})}
				\State $\frameIndex^* = \argmin_\frameIndex{||\viapointMean-\localFrameOrigin||}$  
			\LeftComment{Map via-point $\viapoint_\timeStep$ locally to $p^*$ (see \ref{sec:TP})} 
			\State $\viapoint^{(p^*)} = \{\viapointInput^{(p^*)}, \viapointMean^{(p^*)}, \viapointCovariance^{(p^*)} \}$ \label{algorithm_part:transform_to_local_via-point_start}
			\LeftComment{Update local KMP $p^*$ with via-point} 
			\State{$\kmp^{(p^*)} \gets \kmp^{(p^*)} \cup \viapoint^{(p^*)}$} 
			\EndIf
		\EndFor
		\State{\textbf{Recompute} $\bK^{\local}$, $\bk^*_{\local}$ and $\bk^{**}_{\local}$ for all $p$ where via-points were added (see \ref{sec:KMP})} \label{algorithm_part:transform_to_local_via-point_end}
\end{algorithmic}
	\label{algorithm:local_trajectory_modulation_via_points}
\end{algorithm}

\subsection{Defining and adding local via-points}\label{subsec:via-point_adding}
To incorporate kinesthetic feedback from users during task execution, we introduce via-points at specific inputs and outputs, where corrections are made. Let us assume a time-driven TP-KMP, where $\inputVariable^{(p)} = t$, and $\outputVariable^{(p)} = \bm{x}^{(p)}$ is the end-effector position mapped to frame $p$.
We further assume that the robot tracks a reference $\hat{\outputVariable}_t = \bmu_t$ computed from \eqref{eq:gaussian_product} with a stiffness that is low enough to allow deviations from the desired path through physical interaction.
We propose three different feedback modalities to trigger new via-points:
\begin{itemize}
	\item \textbf{Distance:} via-points are added if the distance between the measured output $\outputVariable_\timeStep$ deviates from the reference $\hat{\outputVariable}_\timeStep$ by a pre-defined threshold $\vert\vert \hat{\outputVariable}_\timeStep - \outputVariable_\timeStep \vert\vert > \threshold_{\xi}$.
	\item \textbf{Force:} via-points are added if the external force applied at the end-effector $\force_t$ exceeds a pre-defined threshold $\vert\vert \force_t \vert\vert > \thresholdForce$.
	\item \textbf{Button press:} via-points can be triggered through a Boolean-type input such as pressing a button, e.g. on a keyboard or on the robot's button interfaces when available. 
\end{itemize}
In all cases, a via-point is defined globally, in a common frame, as $\viapoint_\timeStep = \{\viapointInput = \timeStep, \viapointMean = \outputVariable_{\timeStep}, \viapointCovariance = \threshold_{\Sigma}\bm{I}\}$, where $\threshold_{\Sigma}$ is a small scalar factor.
Different feedback modalities provide users with various options depending on the robot and task.
Force-based adaptation is well-suited for tasks (or sub-tasks) that require little or no contact with the environment, where the robot can interpret external forces as human intention.
A distance criteria is suitable for tasks where the robot has low stiffness and thus would not measure high external forces when perturbed.
In \cref{sec:evaluation}, we showcase the role of various feedback modalities.
When a new global via-point $\viapoint_\timeStep$ is defined, we rely on a proximity-based criteria to determine which KMP to apply it to.
For this, we identify the closest frame to the via-point, ${	\frameIndex^*=\argmin_\frameIndex{||\viapointMean_\timeStep-\localFrameOrigin||}}$.
The via-point is then mapped to the selected frame through ${\bar{\bmu}^{(p^*)} = (\bA^{(p^*)})^{-1}(\bar{\bmu}-\bb^{(p^*)})}$, ${\bar{\bSigma}^{(p^*)} = (\bA^{(p^*)})^{-1}\bar{\bSigma}(\bA^{{(p^*)}^{\top}})^{-1}}$ (see \cref{sec:background} for details) and added to its KMP, entailing the recomputation of kernel-related matrices $\bm{K}_{(p^*)}, \bm{k}^*_{(p^*)}$ and $\bm{k}^{**}_{(p^*)}$. 
The algorithm for interactively adding via-points to local frames is summarized in \cref{algorithm:local_trajectory_modulation_via_points}.

\subsection{Adding new objects through interactive via-point definition}
Our approach permits adding new objects to a skill during runtime, without requiring a new set of demonstrations.
Instead, we leverage \cref{algorithm:local_trajectory_modulation_via_points} to build on an existing TP-KMP, $\tpkmp = \{\localkmp\}^\amountOfFrames_{\frameIndex=1}$, by interactively adding via-points in the frames of new objects.
To achieve this, we employ two key steps.
Firstly, we associate to the new task parameters $\frameOrigin^{(\amountOfFrames+1)},\frameRotationMatrix^{(\amountOfFrames+1)}$ a \textit{placeholder} local KMP with the same inputs as other local KMPs, zero means and high variances, defined by a large scalar $\gamma_D$.
Due to the high variance, the placeholder KMP has a negligible influence in the Gaussian product \eqref{eq:gaussian_product}, not affecting the task unless via-points are added to it.
Next, we use \cref{algorithm:local_trajectory_modulation_via_points} to add via-points to the placeholder KMP, interactively reducing the uncertainty at precise locations, enabling the model to adapt without requiring new demonstrations. 
\cref{algorithm:add_new_frame} summarizes the procedure to add new objects.

\begin{algorithm}[th]
	\caption{\emph{Interactively adding a new object to a TP-KMP}}
	\begin{algorithmic}[1]
\State{\textbf{Input:} TP-KMP $\tpkmp = \{\localkmp\}^\amountOfFrames_{\frameIndex=1}$, variance prior $\threshold_{D}$, new task parameters $\frameOrigin^{(\amountOfFrames+1)},\frameRotationMatrix^{(\amountOfFrames+1)}$.}
\State{\textbf{{Create a \textit{placeholder} local KMP}}}
\Statex{- Define ${\bm{D}^{(\amountOfFrames+1)} =}$ \newline ${\{\inputVariable^{(\amountOfFrames+1)}_\amountOfKMP=\inputVariable^{(\amountOfFrames)}_n, \mean^{(\amountOfFrames+1)}_\kmpIndex = \bm{0}, \covariance^{(\amountOfFrames+1)}_\kmpIndex = \threshold_{D} \bm{I}\}^\amountOfKMP_{\kmpIndex=1}}$}
\Statex{- Define local KMP: ${\tpkmp^{(\amountOfFrames+1)} = \{\frameOrigin^{(\amountOfFrames+1)}, \frameRotationMatrix^{(\amountOfFrames+1)}, \bm{D}^{(\amountOfFrames+1)}\}}$}
\Statex{- Compute $\bK_{(\amountOfFrames+1)}$, $\bk^{*}_{(\amountOfFrames+1)}$ and $\bk^{**}_{(\amountOfFrames+1)}$}
\Statex{- Add new local KMP to TP-KMP: $\tpkmp \gets \tpkmp \cup \tpkmp^{(\amountOfFrames+1)}$, $\amountOfFrames = \amountOfFrames + 1$}
\State{\textbf{Add via-points to new KMP interactively using Algo.\ref{algorithm:local_trajectory_modulation_via_points}}}
\end{algorithmic}
\label{algorithm:add_new_frame}
\end{algorithm}

\subsection{Uncertainty-aware skill extension in regions without demonstrations}
\label{subsec:uncertainties}

Popular variable impedance schemes found in LfD regulate the robot stiffness by the inverse of covariance matrices \cite{medina12, Silverio19}.
When the latter represent the aleatoric uncertainty, they provide an efficient way for robots to be more precise, by being stiffer, where demonstrations showed less variance, following a \textit{minimum intervention control} principle \cite{Todorov04,medina12}.
In the case of epistemic uncertainty, such schemes contribute to better compliance when the robot is uncertain about its actions \cite{Silverio19}.
In kernel-based methods, kernel hyperparameters are typically optimized for the training data, but their choice influences the behavior of the model in regions where data was not shown.
For instance, the kernel length depends on the scale of the input domain, but it also dictates how quickly the epistemic uncertainty increases when moving away from the training data.\footnote{Similar arguments can be made for other hyperparameters such as the noise variance, see \cite{Silverio19}.}
Having ways to clearly distinguish between the two types of uncertainty enables users to better design variable impedance strategies.
%
Similarly to \cite{Kulak21}, the covariance prediction of KMP can be decomposed into two terms, corresponding to aleatoric and epistemic components (see Appendix \ref{sec:kmp_split} for the derivation):
\begin{align}
	& \notag \mathrm{cov}\left[\outputVariable(\inputVariable^*)\right] = \\
	& \underbrace{\bm{k}^{**} - \bm{k}^*\bm{K}^{-1}\bm{k}^{*\top}}_{\bm{\Sigma}_{\mathrm{ep}}^*}+\underbrace{\bm{k}^*\left(\bm{K} + \bm{K}(\lambda_2\bm{\Sigma})^{-1}\bm{K}\right)^{-1}\bm{k}^{*\top}}_{\bm{\Sigma}_{\mathrm{al}}^*}
	\label{eq:kmp_split}
\end{align}
The term $\covariance_\mathrm{ep}^*$ is the same as the variance prediction in Gaussian process regression \cite{Rasmussen05}, corresponding to the epistemic part of the KMP covariance.
The remaining term $\covariance_\mathrm{al}^*$ gives the aleatoric uncertainty.
Using a Cartesian impedance controller, from \eqref{eq:kmp_split}, we propose to compute the robot end-effector stiffness\footnote{We assume a controller $\bm{u}=\bm{G}_P(\hat{\bm{\xi}}-\bm{\xi}) - \bm{G}_D\bm{\dot{\xi}}$, where $\bm{G}_P$, $\bm{G}_D$ are stiffness and damping gains, $\hat{\bm{\xi}}$, $\bm{\xi}$, $\dot{\bm{\xi}}$ are the end-effector desired state, current state and current velocity. $\bm{G}_D$ changes with the joint configuration, to ensure that the system is critically damped in Cartesian space \cite{Ott08}. Redundancy is further addressed at joint level using a null space term that keeps the elbow in a natural position.} using 
\begin{equation}
	\bm{G}_P = w_1 \cdot \left(\delta_\mathrm{ep}\covariance_\mathrm{ep}^{*}\right)^{-1} + w_2 \cdot \left(\delta_\mathrm{al} \covariance_\mathrm{al}^{*}\right)^{-1},
	\label{eq:stiffness_regulation}
\end{equation}
where $w_1\in\left[0,1\right]$ is a sigmoid function that depends on the epistemic uncertainty ${w_1 (\sigma_\mathrm{ep}^2) = 1/(1+e^{-c_1(\sigma_\mathrm{ep}^2-c_2)})}$, with ${\sigma_\mathrm{ep}^2 = \mathrm{tr}(\covariance_\mathrm{ep}^*)/\outputDimension}$, and $w_2 = 1 - w_1$. Through the parameters $c_1>0,c_2$ one is able to regulate the rate at which the robot stiffness switches between being governed by $\covariance_\mathrm{ep}^{*}$ and $\covariance_\mathrm{al}^{*}$, while ensuring continuity in the resulting accelerations.
Parameters $\delta_\mathrm{ep}$, $\delta_\mathrm{al}$ re-scale the uncertainties without modifying the kernel parameters, allowing, for example, $\covariance^*_\mathrm{ep}$ to have a stronger influence in $\bm{G}_P$ far from the training data, which permits a faster increase in robot compliance. 
We leverage \eqref{eq:stiffness_regulation} to facilitate the acquisition of new data outside of the training region through physical interactions, while being optimal in a minimal intervention sense \cite{Todorov04,medina12}, in regions where data was provided.

	\section{Evaluation}
	\label{sec:evaluation}
	\subsection{Evaluation on real robot}\label{subsec:eval_on_robot}
%
We evaluate our approach on a torque-controlled 7-DoF robot in an industrial scenario where an inner ring of a ball bearing needs to be transferred between two boxes (`box 1' and `box 2') placed at different locations on a workbench.\footnote{You can find more information and videos at  \href{https://github.com/DLR-RM/interactive-incremental-learning}{github.com/DLR-RM/interactive-incremental-learning}}
We provide $M=4$ demonstrations with different box positions, 
see \Cref{fig_overview_generalization}-top. 
We use a time-driven representation with $\bs_{h,m}=t_{h,m}/T_m$, where $t$ is a time step, and learn the end-effector position $\bxi_{h,m}=\bm{x}_{h,m}$.
In order to easily re-scale the skill duration, we map all the inputs to the interval $[0,1]$ by dividing them by the duration of each demonstration $T_m$.
The experiments start with $P=2$, with task parameters $\bm{b}^{(p)}, \bm{A}^{(p)}$ representing the box positions and orientations, respectively.
All local KMPs were initialized from GMMs with 12 components, trained on locally projected data, and $N=500$ inputs, equally spread over the input space.
We chose a Mat\'ern kernel ($\nu=5/2$) with length scale $l=0.1$ and noise variance 1.0 (see \cite{Rasmussen05}). Other KMP hyperparameters were $\lambda_1=0.1$, $\lambda_2=1$, $\alpha=1$, chosen empirically.
For completeness, we provide an overview of hyperparameters 
in Appendix \ref{subsec:hyperparam_overview}. 
In all experiments, via-points are added with $\gamma_\Sigma = 1\times 10^{-8}$.
Our algorithm successfully generalizes to novel box positions. 
However, when they deviate significantly from the demonstrated ones, TP-GMM \cite{Calinon16}, as well as the original TP-KMP formulation \cite{Huang19} fail (see \cref{fig_overview_generalization}-bottom).
\subsection{Experiment 1: Improving generalization with interactively-defined local via-points}\label{subsec:experiment1}
\begin{figure}[t]
	\centering
  	\begin{subfigure}[c]{0.45\textwidth}
		\centering
		\includegraphics[width=\columnwidth]{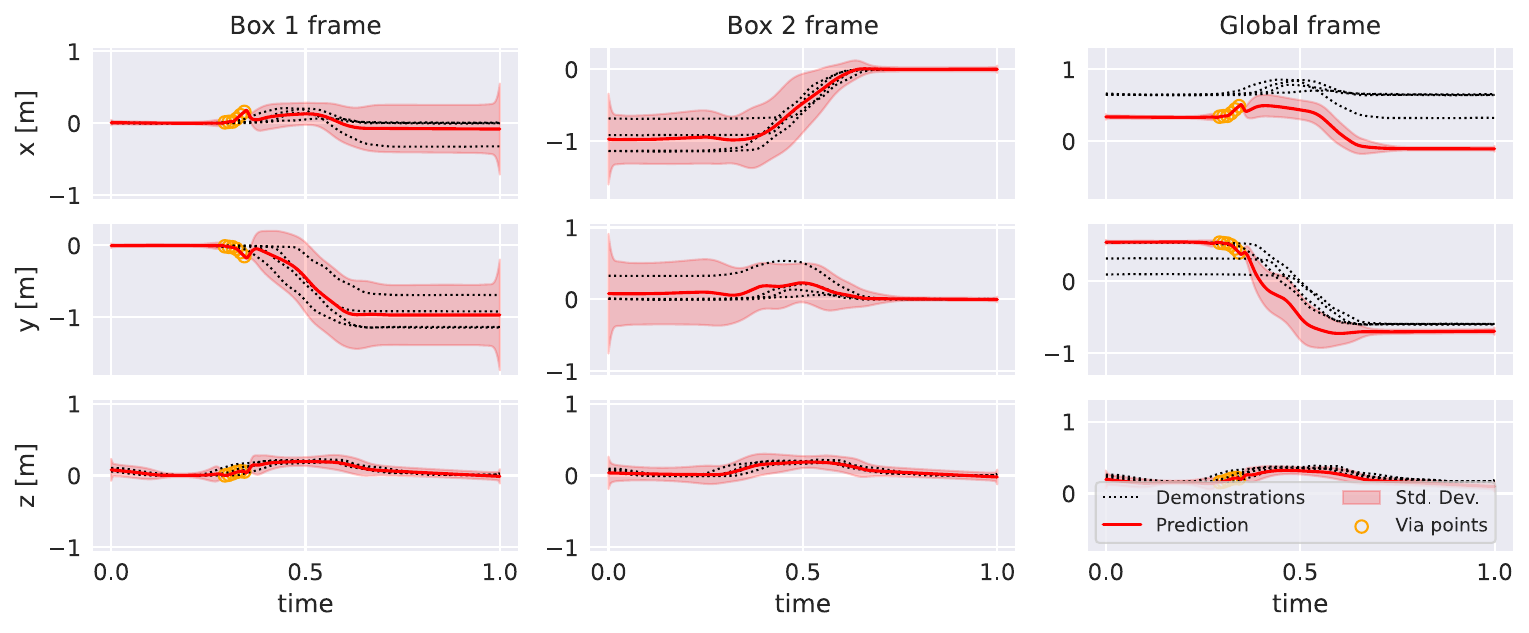}
	\end{subfigure}
	\begin{subfigure}[c]{0.45\textwidth}
		\centering
		\begin{subfigure}[c]{0.4\textwidth}
			\centering
			\includegraphics[width=\textwidth]{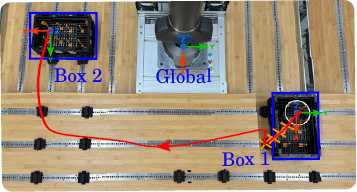}
		\end{subfigure}
		\begin{subfigure}[c]{0.54\textwidth}
			\centering
			\includegraphics[width=\textwidth]{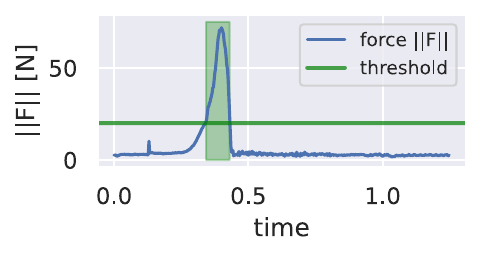}
		\end{subfigure}
	\end{subfigure}
	\caption{Improving generalization with interactively-defined local via-points. Since the robot would fail, a correction is shown close to box 1, indicated by via-points in orange. The threshold for the external force $\thresholdForce$ is set to 20 N. \label{fig_evaluation_experiment1}}
\end{figure}

Since all demonstrations are given at the same height, the model has a high confidence for the $z$ coordinate in all frames, which leads to poor generalization when one of the boxes is moved to a different height (\cref{fig_overview_generalization}-bottom).
If box 2 is moved to a lower height, the trade-off found by the model, from the expected vertical motions in the two frames, is not high enough to successfully move the ring out of box 1, leading to a collision.
Using \cref{algorithm:local_trajectory_modulation_via_points}, we have the ability to directly correct the robot and set via-points based on corrections made online.
In this experiment we used an external force trigger, with threshold $\thresholdForce = 20 N$.
Figure \ref{fig_evaluation_experiment1} shows the results obtained for this experiment.
With the robot running the model, a user applies a force at the robot end-effector, helping it avoid the collision and successfully move the ring out of the box.
Via-points are added to the frame of box 1 and, through $\bm{b}^{(1)},\bm{A}^{(1)}$, are mapped to the global frame.
With the added via-points, the robot achieves a success rate of $93\%$ across 15 box configurations (with three different heights), improved over the $60\%$ success rate of TP-GMM \cite{Calinon16} and original TP-KMP \cite{Huang19} baselines (see Additional Material \ref{appendix:evaluation_details}). 

\subsection{Experiment 2: Interactively adding a new object to skill}
\begin{figure}
	\centering
	\begin{subfigure}[c]{0.32\textwidth}
		\includegraphics[width=\columnwidth]{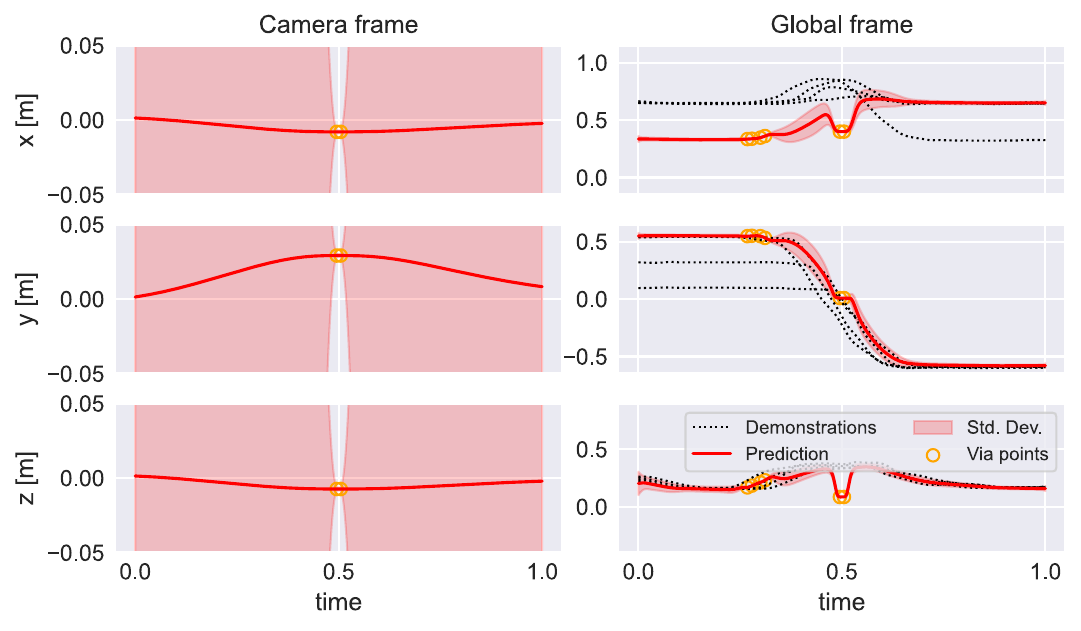}
	\end{subfigure}
	\begin{subfigure}[c]{0.16\textwidth}
		\includegraphics[width=.9\columnwidth]{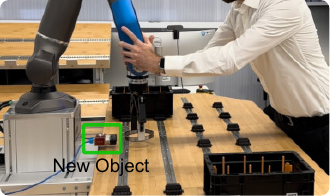}\\
		\includegraphics[width=.9\columnwidth]{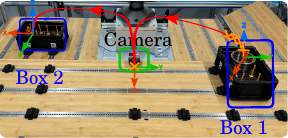}
	\end{subfigure}
	\caption{Interactively adding a new object to existing skill. We add a camera to the scene, associate it with a new frame and add via-points to pass in front of the camera, this time by pressing a button.} 
	\label{fig_adding_camera_frame_inbetween}
\end{figure}
We introduce a new challenge by requiring the robot to inspect the ring using a camera, before placing it in box 2.
For this, the user increments the existing skill by interactively adding via-points to a new frame, given by the camera pose, using Algorithm \ref{algorithm:add_new_frame}.
The new via-points adapt the robot's trajectory to pass in front of the camera before placing the ring in box 2.
In this experiment we use the button interfaces in the robot's last link to trigger new via-points through button presses.
We further use $P=3$, due to the new frame, and $\threshold_D=1\times 10^4$.
The experimental results are shown in \cref{fig_adding_camera_frame_inbetween}. 
With the task running, the user adds new via-points near the camera (top-right), reducing the uncertainty in the camera frame. 
Once the model is updated with the new via-points, new executions pass in front of the camera (bottom-right).

\subsection{Experiment 3: Incremental learning at non-demonstrated inputs}
We further add a camera to the robot end-effector, which validates the correct insertion of the ring in box 2.
This requires the robot to move up after placing the ring -- a skill extension to a set of inputs that were not shown in the demonstrations (${t > 1.0}$).
We leverage \eqref{eq:stiffness_regulation} to ensure that the robot becomes compliant by swiftly lowering its stiffness in response to the increase in epistemic uncertainty.

\begin{figure}[h]
	\centering
	\begin{subfigure}[c]{0.20\textwidth}
		\centering
		\vspace{0.2cm}
		\includegraphics[width=0.9\columnwidth]{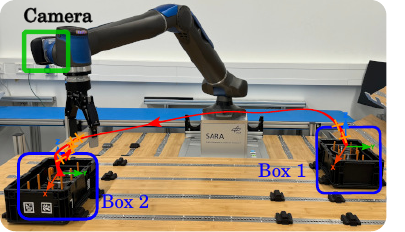}
		\includegraphics[width=\columnwidth]{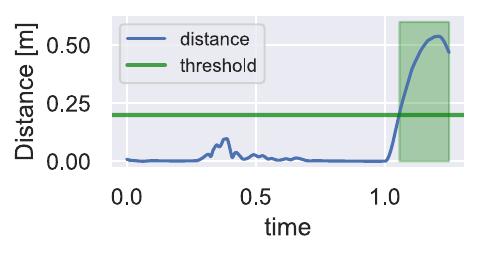}
	\end{subfigure}
	\begin{subfigure}[c]{0.28\textwidth}
		\vspace{-0.1cm}
		\includegraphics[width=\columnwidth]{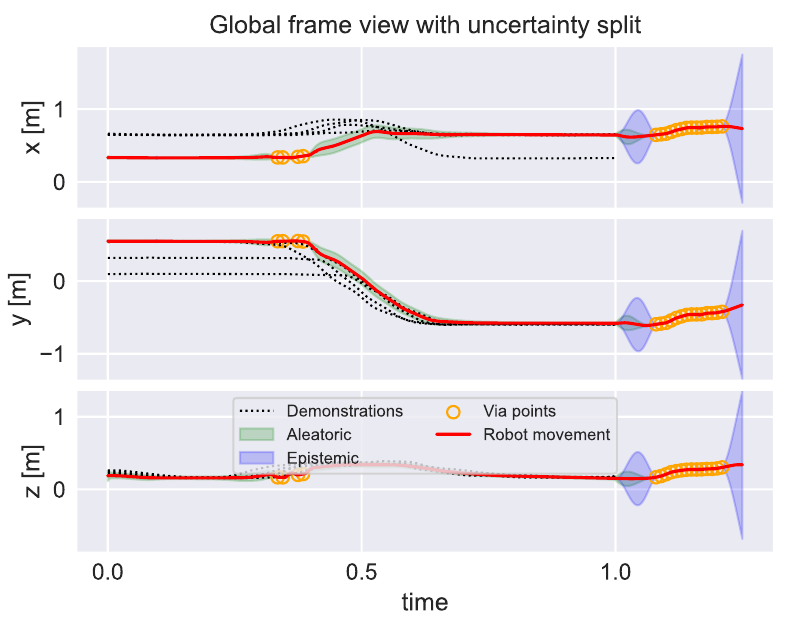}
	\end{subfigure}
		\caption{Adding via-points at $t>1.0$, triggered by distance threshold $\thresholdTrajectory = 20 m$ leads to increase of uncertainty outside of demonstrated area.
			\label{fig_adding_frame_before_demo}}
	\end{figure}

We set $c_1=5\times 10^3	$, $c_2=1.5\times 10^{-3}$, $\delta_\mathrm{ep}=1\times 10^3$, $\delta_\mathrm{al}=1$, chosen empirically.
\begin{figure}
	\centering
	\begin{subfigure}[c]{0.50\textwidth}
		\includegraphics[width=\textwidth]{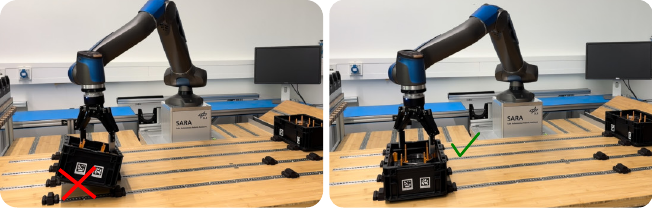}\\
	\end{subfigure}
	\hfill
	\begin{subfigure}[c]{0.50\textwidth}
		\centering
		\includegraphics[width=0.49\columnwidth]{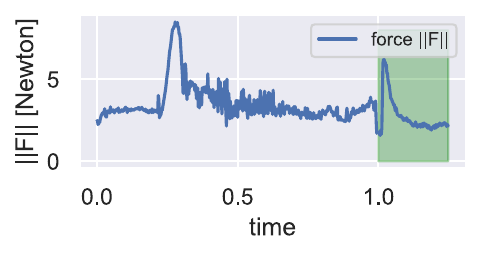}
		\includegraphics[width=0.49\columnwidth]{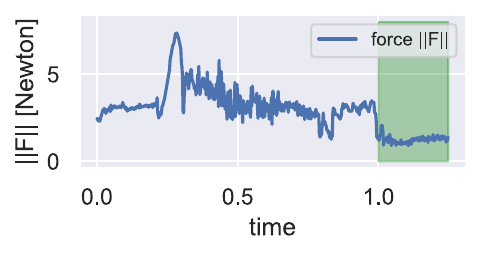}
	\end{subfigure}
		\caption{\textbf{Left:} Computing stiffness gains from \eqref{eq:kmp_cov} leads to increased interaction forces at non-demonstrated inputs. \textbf{Right:} Using our approach \eqref{eq:kmp_split}--\eqref{eq:stiffness_regulation} the robot reacts quickly to the increase in the epistemic uncertainty, lowering its stiffness. $||\mathbf{F}||$ represents the Euclidean norm of vector $\mathbf{F}$.} \label{fig_evaluaton_figure_for_adding_frame_at_the_end}
	\end{figure}
We consider a baseline where the gains are computed from the original KMP covariance \eqref{eq:kmp_cov} as $\bm{G}_P = (\mathrm{cov}\left[\outputVariable(\inputVariable^*)\right])^{-1}$.
To address numerical issues and ensure a maximum stiffness value we regularize $\mathrm{cov}\left[\outputVariable(\inputVariable^*)\right]$, $\delta_\mathrm{al}\bSigma^*_\mathrm{al}$, $\delta_\mathrm{ep}\bSigma^*_\mathrm{ep}$ with a small scalar $1.5\times 10^{-3}$.
Figure \ref{fig_evaluaton_figure_for_adding_frame_at_the_end} shows that using our approach \eqref{eq:stiffness_regulation} the interaction forces after ${t=1.0}$ are negligible.
Note that far from the training data, the KMP expectation \eqref{eq:kmp_mean} converges to zero, as we assume a zero mean prior (similarly to GPs \cite{Rasmussen05}).
With the improved compliance behaviors introduced by \eqref{eq:stiffness_regulation}, the user is able to add via-points such that the robot moves up after the ring insertion.
In this experiment we used a distance threshold with $\thresholdTrajectory = 0.2 m$, taking advantage of the reference trajectory going to zero to trigger the definition of via-points. 
The bottom-left image in \cref{fig_adding_frame_before_demo} shows the increase in distance as via-points are added, while the plots on the right show the resulting robot motion and uncertainties.
Appendix \ref{sec:stiffness_complete} shows the detailed stiffness profiles from Experiment 3 including a sensitivity analysis of the kernel length scales. 

\begin{table}[]
	\begin{center}
		\caption{Mean $\pm$ standard deviation distance from ground truth for the pick-and-place task from Yao \etal \cite{Yao24}. All results besides TP-KMP \cite{Huang19} and our approach are generated using the data and code provided by \cite{Yao24}. We added via-points to the start and stop positions.}
		\label{tab:comparison_approaches_tppromp}
		\resizebox{\columnwidth}{!}{%
			\begin{tabular}{c|c|c|c}
				& Start (mm) & End (mm) & Average (mm) \\
				\hline\hline
				ProMP \cite{Paraschos13} & 107.78 $\pm$ 7.99 & 137.84 $\pm$ 18.10 & 250.68 $\pm$ 137.13 \\
				TP-GMM (time-based) \cite{Calinon16} & 28.22 $\pm$ 14.90 & 74.11 $\pm$ 48.06 & 58.76 $\pm$ 21.43\\ 
				TP-GMM (dynamic) \cite{Calinon16} & 27.54 $\pm$ 15.56 & 129.42 $\pm$ 100.18 & 83.45 $\pm$ 43.01 \\ 
				KMP \cite{Huang19} & 25.80 $\pm$ 11.39 & 73.99 $\pm$ 36.44 & 57.48 $\pm$ 21.62\\
				TP-ProMP \cite{Yao24} & 32.62 $\pm$ 11.39 & 48.60 $\pm$ 20.68 & \bm{$48.51 \pm 12.21$}\\
				TP-KMP \cite{Huang19} & 39.30 $\pm$ 29.06 & 6.93 $\pm$ 5.08 & 51.95 $\pm$ 26.44\\ 
				\hline
				\multirow{2.5}{*}{\shortstack{\textbf{Our approach} \\ \textbf{(with via-points)}}} & \multirow{2.5}{*}{\shortstack{\bm{$3.88 \times 10^{-4}$} \\ \bm{$\pm 2.54 \times 10^{-4}$}}} & \multirow{2.5}{*}{\shortstack{\bm{$1.16\times 10^{-3}$} \\ \bm{$\pm 6.38 \times 10^{-4}$}}} & \multirow{2.5}{*}{\shortstack{50.23 $\pm$ 25.58}} \\
				&&&
			\end{tabular} 
		}
	\end{center}
\end{table}

\subsection{Evaluation on toy example}\label{subsec:eval_on_toy_example}
In \cref{tab:comparison_approaches_tppromp} we compare the performance of our approach in \ref{subsec:interactive_vp} to other LfD approaches (with a strong focus on task-parameterized formulations).
For this comparison, we follow the evaluation framework of Yao et al. \cite{Yao24} and ran both TP-KMP \cite{Huang19} and our interactive approach on the pick-and-place task; the results are based on 100 runs with different start and end points.
Our approach showed significant improvement at both the start and end points, delivering the best results in this comparison for start and end precision.
This is due to showed corrections at the start and end-point, which we did by locally adding via-points at the origin of the corresponding reference frames.
Our approach also compares favorably in terms of average precision, despite not including other via-points than at start and end, similarly to \cite{Yao24}. 

	\section{Discussion}
	\label{sec:discussion}
\subsection{Analysis of the results}
\cref{fig_evaluation_experiment1} shows that in Experiment 1 an external force trigger successfully allows for the definition of via-points in the nearest frame (that of box 1).
Thanks to a small via-point variance, the via-points are mapped to the global frame by \eqref{eq:gaussian_product} improving generalization quality by avoiding a collision.
\cref{fig_adding_camera_frame_inbetween}, Experiment 2 illustrates how the definition of a placeholder frame with large variance can be used in combination with the via-point insertion mechanism from Experiment 1 to introduce behaviors with respect to objects that were not present in the demonstrations.
Particularly, one observes that the variance in frame 3 decreases after the via-points are added, with the latter being successfully mapped to the global frame.
Finally, Experiment 3 shows how our stiffness regulation approach leverages the epistemic uncertainty to enhance the robot's compliance beyond the initial set of demonstrations.
While one could argue that a similar effect could be achieved by manually lowering the stiffness at $t>1.0$, this would require manually keeping track of the duration of demonstrations, as well as the exact locations of newly added via-points.
Our approach automates this by relying on the data properties, increasing the epistemic uncertainty both before and after via-points (\Cref{fig_adding_frame_before_demo}-right).

\subsection{Limitations}
In our approach, via-points are added in all Cartesian DoFs, even though a correction might only be required in a subset thereof (\eg height in the first experiment).
Since KMPs allow the definition of via-point covariances with different diagonal entries, one can, in principle, selectively set higher precision only on the DoFs that receive a corrective action (keeping the others as in the training data).
This, however, requires a more complex interaction mechanism which extracts the user intention, a topic that we plan to address in future research.
Another possible limitation is that our approach currently does not allow for combining new via-points with existing ones at the same location.
If a via-point is added at a location where one already existed, the only way to account for the newly added information is to replace the pre-existing via-point.
To provide users with more options for skill re-use, in future work we will investigate mechanisms to interactively remove via-points.

	\section{Conclusion}
	\label{sec:conclusion}
	We presented an interactive imitation learning framework that leverages both local and global modulations of trajectory distributions to address the problem of generalization in LfD. 
To improve the generalization quality and incrementally add new features to a demonstrated skill, the framework allows the interactive definition of local via-points.
This is facilitated by a variable impedance scheme that leverages epistemic uncertainties to augment skills beyond the demonstrations. 
Our results, evaluated on a ring-loading task using a torque-controlled, 7-DoF robot, show that our framework permits users to incrementally build on an initial model of a skill by interactively correcting errors and adding new behaviors in any phase of the task.
This work has significant implications for the development of robots that can learn from demonstration and generalize their skills to new situations, making them more versatile and effective in real-world applications.

	\section*{Acknowledgments}
	\label{sec:acknowledgements}
	The authors thank Thomas Eiband and Korbinian Nottensteiner for their discussions.
	
	\appendix
	\section{Additional material}
	\subsubsection{Decomposition of KMP covariance into a sum of epistemic and aleatoric terms}
\label{sec:kmp_split}

We here show that, similarly to \cite{Kulak21}, the covariance prediction of KMP can be decomposed into two distinct terms, corresponding to aleatoric and epistemic components.
Using the Woodbury identity ${(\bm{A}+\bm{U}\bm{B}\bm{V})^{-1} = \bm{A}^{-1} - \bm{A}^{-1}\bm{U}\left(\bm{B}^{-1}+\bm{V}\bm{A}^{-1}\bm{U}\right)^{-1}\bm{V}\bm{A}^{-1}}$ with $\bm{V}=\bm{U}=\bm{I}$, we can re-write \eqref{eq:kmp_cov} as:
\begin{align}
	&\notag \mathrm{cov}\left[\outputVariable(\inputVariable^*)\right] = \\
	& \notag =\bm{k}^{**} - \bm{k}^*\left[\bm{K}^{-1}-\bm{K}^{-1}\left(\bm{K}^{-1} + (\lambda_2\bm{\Sigma})^{-1}\right)^{-1}\bm{K}^{-1}\right]\bm{k}^{*\top}\\
	& \notag = \bm{k}^{**} - \bm{k}^*\bm{K}^{-1}\bm{k}^{*\top}+\bm{k}^*\bm{K}^{-1}\left(\bm{K}^{-1} + (\lambda_2\bm{\Sigma})^{-1}\right)^{-1}\bm{K}^{-1}\bm{k}^{*\top}\\
	& \notag = \underbrace{\bm{k}^{**} - \bm{k}^*\bm{K}^{-1}\bm{k}^{*\top}}_{\bm{\Sigma}_{\mathrm{ep}}^*}+\underbrace{\bm{k}^*\left(\bm{K} +  \bm{K}(\lambda_2\bm{\Sigma})^{-1}\bm{K}\right)^{-1}\bm{k}^{*\top}}_{\bm{\Sigma}_{\mathrm{al}}^*},
\end{align}
where, in the last step, we used ${\bm{A}^{-1}\left(\bm{A}^{-1}+\bm{B}^{-1}\right)^{-1}\bm{A}^{-1} = \left(\bm{A}+\bm{A}\bm{B}^{-1}\bm{A}\right)^{-1}}$ (omitting $\alpha$ for the sake of the derivation).
Note the clear separation between the two terms in \eqref{eq:kmp_split}.
The term $\covariance_\mathrm{ep}^*$ is the same as the variance prediction in Gaussian process regression \cite{Rasmussen05}, corresponding to the epistemic part of the KMP covariance.
The remaining term $\covariance_\mathrm{al}^*$ gives the aleatoric uncertainty.

\label{sec:appendix_uncertainties}

\subsubsection{Hyper-parameter overview}
\label{subsec:hyperparam_overview}
\begin{table}[]
	\begin{center}
		\caption{Hyper-parameter overview}
		\label{tab:hyperparameter_overview}
		\resizebox{\columnwidth}{!}{%
			\begin{tabular}{l|c|c|c}
				Hyper-parameter & \multirow{2}{*}{\shortstack{Used in\\paper}} & \multirow{2}{*}{\shortstack{Validated\\range/options}} & Remark \\ [2ex]
				\hline\hline
				Nr. of Gaussians to initialize GMM & 12 & [7; 40] & $\uparrow\! \mathbb{N}$ : for more complex/longer tasks\\ 
				Regularization term KMP mean $\lambda_1$ & 0.1 & [0.1; 1.0] & $\uparrow\! \lambda_1$: closer following the mean distribution of the KMP\\
				Regularization term KMP variance $\lambda_2$ & 1.0 & [0.1; 1.0] &$\uparrow\! \lambda_2$: closer following the variance distribution of the KMP\\
				Scaling factor for KMP covariance $\alpha$ & 1.0 & [0.1; 1.0] & - \\
				Length scale of kernel $l$ & 0.1 & [0.1; 0.7] & $\uparrow\! l$: for smoother predictions, but less accuracte \\
				Kernel function $k(.,.)$ & matern2 & matern2 / rbf & - \\
				Sigmoid function parameter $c_1$ & 5 $\times 10^{3}$ & [1; $1 \times 10^{4}$] & $\uparrow\! c_1$: control gains switch more quickly\\
				&&&between aleatoric and epistemic uncertainty\\
				Sigmoid function parameter $c_2$ & $1.5 \times 10^{-3}$ & [$1 \times 10^{-4}$; 1] & $\uparrow\! c_2$: epistemic uncertainty takes over for larger values of $\sigma^2_\textrm{ep}$\\
				Epistemic scaling factor $\delta_\mathrm{ep}$ & $1 \times 10^{3}$ & [1;  $1 \times 10^{4}$] & $\uparrow\! \delta_\mathrm{ep}$: increases the influence of epistemic uncertainty\\
				Aleatoric scaling factor $\delta_\mathrm{al}$ & 1 & [1; $1 \times 10^{4}$] & $\uparrow\! \delta_\mathrm{al}$: increases the influence of aleatoric uncertainty\\
			\end{tabular} 
		}
	\end{center}
\end{table}
Table \ref{tab:hyperparameter_overview} shows the different hyperparameters used in this paper and a tested range of values to provide a guideline for choosing hyperparameters.

\subsubsection{Detailed stiffness profiles for Experiment 3 with kernel length scale sensitivity analysis}
\label{sec:stiffness_complete}

\begin{figure}[h]
	\centering
	\includegraphics[width=0.49\columnwidth]{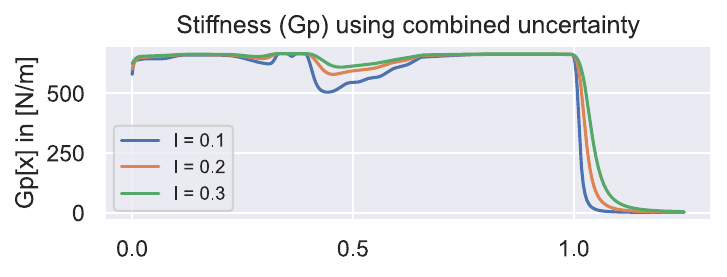}
	\includegraphics[width=0.49\columnwidth]{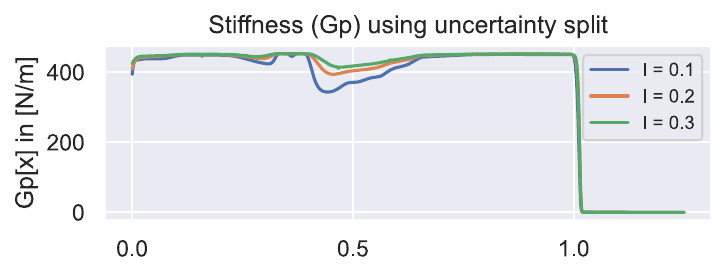}
	\caption{Stiffness values $\bm{G}_P$ for the $x$ dimension, before via-points are added. \textbf{Left:} Stiffness computed using the original KMP covariance \eqref{eq:kmp_cov}, through $\bm{G}_P = (\mathrm{cov}\left[\outputVariable(\inputVariable^*)\right])^{-1}$. \textbf{Right:} Stiffness computed using our approach that separates epistemic and aleatoric uncertainties \eqref{eq:stiffness_regulation}. The different line colors correspond to different kernel lengths $l$, which we compared.
		\label{fig_stiffness_no_via}}
\end{figure}

\begin{figure}[h]
	\centering
	\includegraphics[width=0.49\columnwidth]{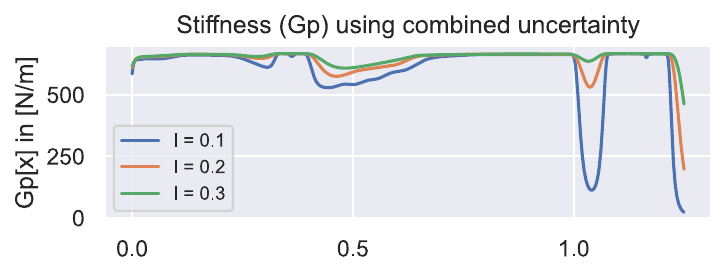}
	\includegraphics[width=0.49\columnwidth]{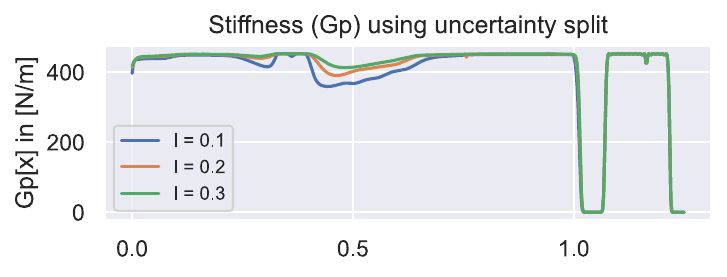}
	\caption{Stiffness values $\bm{G}_P$ for the $x$ dimension, after via-points are added. Left and right plots show the curves obtained with uncertainties computed from \eqref{eq:kmp_cov} and \eqref{eq:kmp_split}, respectively, similarly to \Cref{fig_stiffness_no_via}, using different kernel length values $l$.
		Our approach (using uncertainty split) successfully makes the robot compliant in regions without data.
		\label{fig_stiffness_via}}
\end{figure}

Figures \ref{fig_stiffness_no_via}--\ref{fig_stiffness_via} give a detailed view of the stiffness profiles which we observed in Experiment 3 (we focus on the $x$ dimension for simplicity).

In \Cref{fig_stiffness_no_via} we see the stiffness profiles before any via-points are added at $t>1.0$.
The left plot shows that after the skill ends at $t=1.0$, the stiffness computed by inverting the covariance \eqref{eq:kmp_cov} decreases slowly to zero, an effect that is more noticeable as the kernel length $l$ increases.
This behavior explains the increased force profiles that we observed in \Cref{fig_evaluaton_figure_for_adding_frame_at_the_end}, bottom left, since the robot collides with the box for a few instants as the stiffness approaches zero, making it unsafe to interact with.
The right plot shows how our proposed approach \eqref{eq:stiffness_regulation} makes the stiffness approach zero faster and with lower sensitivity to the value of $l$. 
The parameter $c_2$ in the activation function allows regulation of the epistemic uncertainty threshold at which $\bm{G}_P$ transitions from being governed by aleatoric uncertainty to epistemic uncertainty.
For this reason, even minor increases in epistemic uncertainty are sufficient to trigger a decrease in stiffness making the robot compliant and resulting in the lower interaction forces seen in \Cref{fig_evaluaton_figure_for_adding_frame_at_the_end}, bottom right.
Due to the continuity of the sigmoid function, the transition happens without discontinuities in the generated control actions.
Figure \ref{fig_stiffness_via} shows that after via-points are added, our approach successfully makes the robot compliant in regions without data, regardless of the kernel length scale, demonstrating improved compliance and efficiency.
	
	\bibliographystyle{IEEEtran}
	\bibliography{bibliography.bib}
	
	\clearpage
	\setcounter{page}{1}
	\renewcommand*{\thepage}{A\arabic{page}}
	\onecolumn
	\markboth{Knauer \MakeLowercase{\textit{et al.}}: Interactive incremental learning - Additional material}
	{Knauer \MakeLowercase{\textit{et al.}}: Interactive incremental learning - Additional material} 
	\section{Additional material}

\subsection{Key notations}
\label{sec:notations}

\cref{tab:notation} summarizes the key notations used in our framework.
\begin{table*}[]\caption{Description of key notations}
	\centering 
	\begin{tabular}{l l p{0.58\textwidth} }
		\toprule
		\rowcolor{Light1}
		$\inputDimension  \in \mathbb{N}$ & $\triangleq$ & Input dimension \\
		\rowcolor{Light1}
		$\outputDimension \in \mathbb{N}$ & $\triangleq$ & Output dimension \\
		\rowcolor{Light1}
		$\amountOfDatapoints \in \mathbb{N}$ & $\triangleq$ & Number of demonstrations \\
		\rowcolor{Light1}
		$H  \in \mathbb{N}$ & $\triangleq$ & Number of data points per demonstration\\
		\rowcolor{Light1}
		$\amountOfKMP \in \mathbb{N}$ & $\triangleq$ & Number of gaussians per KMP\\
		\rowcolor{Light1}
		$\inputVariable\in\mathbb{R}^\mathcal{I}$ & $\triangleq$ & Input variable \\
		\rowcolor{Light1}
		$\outputVariable \in \mathbb{R}^\mathcal{O}$ & $\triangleq$ & Output variable \\
		\rowcolor{Light1}
		$\{\{\inputVariable_{h,m},\outputVariable_{h,m}\}_{h=1}^H\}_{m=1}^M$ & $\triangleq$ & Set of demonstrations \\
		\rowcolor{Light1}
		$\bm{x}$ & $\triangleq$ & End-effector position in Cartesian space\\
		
		\rowcolor{Light2}
		$\amountOfFrames \in \mathbb{N}$ & $\triangleq$ & Number of frames in a TP-KMP\\
		\rowcolor{Light2}
		$\frameIndex = 1,\ldots,P$ & $\triangleq$ & Frame index \\
		\rowcolor{Light2}
		$\frameIndex^*$ & $\triangleq$ & closest frame to a given via-point \\
		\rowcolor{Light2}
		$\frameOrigin^{(\frameIndex)}$, $\frameRotationMatrix^{(\frameIndex)}$ & $\triangleq$ & Task parameters of frame $p$ \\
		\rowcolor{Light2}
		$\bs^{\local}, \bxi^{\local}$ & $\triangleq$ & Demonstrations represented locally in frame $p$ \\	
		\rowcolor{Light2}
		$\localMean$, $\localCovariance$ & $\triangleq$ & Local mean and covariance of frame $p$ \\
		\rowcolor{Light2}
		$\hat{\mean}^{\local}$, $\hat{\covariance}^{\local}$ & $\triangleq$ & Mean and covariance of frame $p$ in global frame \\
		\rowcolor{Light2}
		$\globalViapoint = \{\bar{s}, \viapointMean, \viapointCovariance \}$ & $\triangleq$ & Via-point in global frame \\
		\rowcolor{Light2}
		$\localViapoint = \{\bar{s}^{(p)}, \localViapointMean, \localViapointCovariance \}$ & $\triangleq$ & Via-point in local frame $p$ \\
		
		\rowcolor{Light3}
		$\localkmpVanilla = \{ \localkmpInput, \localkmpMean, \localkmpCovariance \}^\amountOfKMP_{\kmpIndex=1}$ & $\triangleq$ & Local reference trajectory distribution \\
		\rowcolor{Light3}
		$\localkmp = \{\localFrameOrigin, \localFrameRotationMatrix, \localkmpVanilla \}$ & $\triangleq$ & Local KMP \\
		\rowcolor{Light3}
		$\tpkmp = \{\localkmp\}^\amountOfFrames_{\frameIndex=1}$ & $\triangleq$ & TP-KMP \\
		\rowcolor{Light3}
		$\lambda_1$, $\lambda_2$ & $\triangleq$ & Regularization terms for KMP mean and covariance \\
		\rowcolor{Light3}
		$\alpha$ & $\triangleq$ & Scaling factor for KMP covariance \\
		\rowcolor{Light3}
		$l$ & $\triangleq$ & Length scale of the kernel\\
		\rowcolor{Light3}
		$k(.,.)$ & $\triangleq$ & Kernel function \\
		\rowcolor{Light3}
		$\bK$, $\bk^*$, $\bk^{**}$ & $\triangleq$ & Kernel matrices for different combinations of inputs \\
		\rowcolor{Light3}
		$\bK_{(p)}$, $\bk^*_{(p)}$, $\bk^{**}_{(p)}$ & $\triangleq$ & $p~\mathrm{th}$ local KMP kernel matrices \\
		\rowcolor{Light3}
		$\bSigma^*_\mathrm{ep}$, $\bSigma^*_\mathrm{al}$ & $\triangleq$ & Epistemic and aleatoric terms of KMP covariance \\
		\rowcolor{Light3}
		$\sigma^2_\mathrm{ep}$ & $\triangleq$ & Epistemic variance (diagonal element of $\bSigma^*_\mathrm{ep}$) \\
		
		\rowcolor{Light4}
		$\force$ & $\triangleq$ & External force measured at the end-effector \\
		\rowcolor{Light4}
		$\thresholdForce$ & $\triangleq$ & Threshold for triggering force-based via-points\\
		\rowcolor{Light4}
		$\thresholdTrajectory$ & $\triangleq$ & Threshold for triggering trajectory-based via-points \\
		\rowcolor{Light4}
		$\gamma_\Sigma$ & $\triangleq$ & Prior covariance of new via-points \\
		\rowcolor{Light4}
		$\gamma_D$ & $\triangleq$ & Prior covariance of \textit{placeholder} KMP \\
		\rowcolor{Light4}
		$w_1$, $w_2$ & $\triangleq$ & Weights of epistemic and aleatoric uncertainty \\
		\rowcolor{Light4}
		$c_1$, $c_2$ & $\triangleq$ & Sigmoid function parameters \\
		\rowcolor{Light4}
		$\delta_\mathrm{ep}$, $\delta_\mathrm{al}$ & $\triangleq$ & Epistemic and aleatoric scaling factors \\
		\rowcolor{Light4}
		$\bm{G}_P$, $\bm{G}_D$ & $\triangleq$ & Stiffness and damping gains \\
		\bottomrule
	\end{tabular}
	\label{tab:notation}
\end{table*}

\subsection{Key acronyms}
\label{sec:acronyms}

\cref{tab:glossar} shows a glossary of important acronyms in our approach.
\begin{table*}[]\caption{Glossary of important acronyms}
	\centering 
	\begin{tabular}{l l p{0.58\textwidth} }
		\toprule
		LfD & $\triangleq$ & Learning from Demonstration \\
		DoF & $\triangleq$ & Degrees of Freedom \\
		TP & $\triangleq$ & Task Parameterization \\
		(TP)-GMM & $\triangleq$ & (Task parameterized)- Gaussian Mixture Model \cite{Calinon16}\\
		(TP)-GMR & $\triangleq$ & (Task parameterized)- Gaussian Mixture Regression \cite{Calinon16}\\
		(TP)-KMP & $\triangleq$ & (Task parameterized)- Kernelized Movement Primitive \cite{Huang19}\\
		DMP & $\triangleq$ & Dynamic Movement Primitives \cite{Ijspeert13}\\
		ProMP & $\triangleq$ &  Probabilistic Movement Primitives \cite{Paraschos13}\\
		LfEC & $\triangleq$ & Learning from Extrapolated Corrections \cite{Zhang2019}\\
		LfI & $\triangleq$ & Learning from Interventions \cite{spencer2020} \\ 
		\bottomrule
	\end{tabular}
	\label{tab:glossar}
\end{table*}

\subsection{Evaluation details}\label{appendix:evaluation_details}
\cref{tab:quantitative_comparison_approaches} and \cref{tab:quantitative_comparison_approaches_2} show more details of the quantitative evaluation of 15 different task-parameterized scenarios, mentioned in \cref{subsec:experiment1} of the paper.

\newcolumntype{M}[1]{>{\centering\arraybackslash}m{#1}}
\begin{table*}[hb!]
	\begin{center}
		\centering
		\caption{(PART I) Quantitative evaluation of 15 different task parameterized scenarios.
			\protect\greencheckmark successful, \protect\yellowcheckmark successful but contact force was above 10 N, \protect\xSymbol failure,  \protect\yellowxSymbol* successful but only after correcting the whole trajectory, which we count as failure.}
		\label{tab:quantitative_comparison_approaches}
		\begin{tabular}{M{1cm}M{6cm}M{2.4cm}M{2.4cm}M{2.4cm}}
			\toprule[0.12em]
			\# & Configuration & TP-GMM & \textit{vanilla} TP-KMP & Ours \\
			\midrule
			1 & \includegraphics[width=0.3\textwidth]{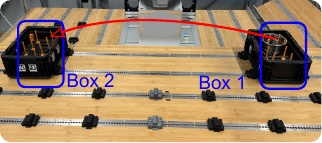}& \yellowcheckmark & \yellowcheckmark & \greencheckmark\\
			2 & \includegraphics[width=0.3\textwidth]{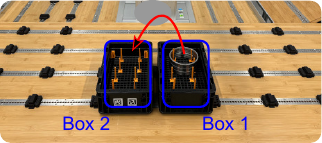}& \yellowcheckmark & \yellowcheckmark & \greencheckmark \\
			3 & \includegraphics[width=0.3\textwidth]{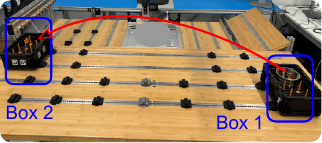}& \greencheckmark & \greencheckmark & \greencheckmark \\
			4 & \includegraphics[width=0.3\textwidth]{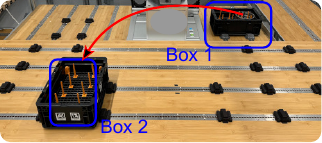}& \xSymbol& \xSymbol& \greencheckmark \\
			5 & \includegraphics[width=0.3\textwidth]{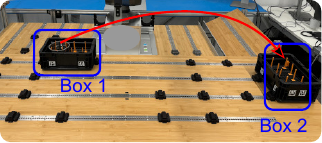}& \xSymbol& \xSymbol& \yellowxSymbol*\\
			6 & \includegraphics[width=0.3\textwidth]{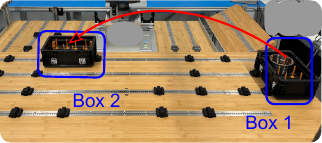}& \xSymbol& \xSymbol& \greencheckmark \\
			7 & \includegraphics[width=0.3\textwidth]{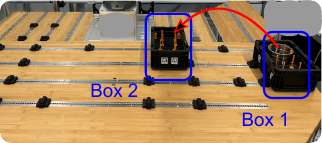}& \yellowcheckmark & \yellowcheckmark & \greencheckmark \\
			8 & \includegraphics[width=0.3\textwidth]{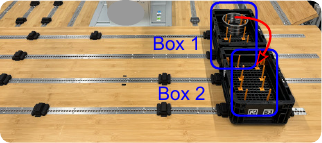}& \greencheckmark & \greencheckmark & \greencheckmark \\
			\multirow{5}{*}{continues in \cref{tab:quantitative_comparison_approaches_2}}
		\end{tabular}
	\end{center}
\end{table*}

\begin{table*}[]
	\begin{center}
		\centering
		\caption{(PART II) Quantitative evaluation of 15 different task parameterized scenarios.
			\protect\greencheckmark successful, \protect\yellowcheckmark successful but contact force was above 10 N, \protect\xSymbol failure,  \protect\yellowxSymbol* successful but only after correcting the whole trajectory, which we count as failure.}
		\label{tab:quantitative_comparison_approaches_2}
		\begin{tabular}{M{1cm}M{6cm}M{2.4cm}M{2.4cm}M{2.4cm}}
			\toprule[0.12em]
			\# & Configuration & TP-GMM & \textit{vanilla} TP-KMP & Ours \\
			\midrule
			9 & \includegraphics[width=0.3\textwidth]{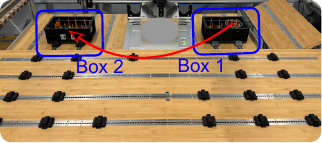}& \xSymbol& \xSymbol& \greencheckmark \\
			10 & \includegraphics[width=0.3\textwidth]{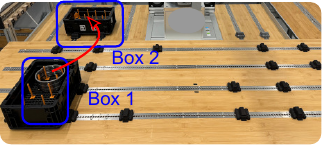}& \xSymbol& \xSymbol& \greencheckmark \\
			11 & \includegraphics[width=0.3\textwidth]{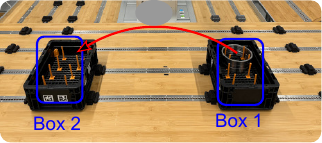}& \greencheckmark & \greencheckmark & \greencheckmark \\
			12 & \includegraphics[width=0.3\textwidth]{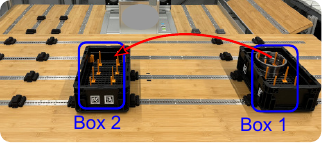}& \greencheckmark & \greencheckmark & \greencheckmark \\
			13 & \includegraphics[width=0.3\textwidth]{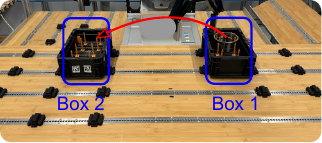}& \greencheckmark & \greencheckmark & \greencheckmark \\
			14 & \includegraphics[width=0.3\textwidth]{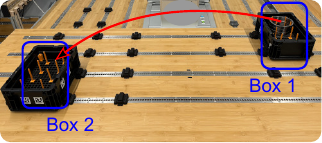}& \yellowcheckmark & \yellowcheckmark & \greencheckmark \\
			15 & \includegraphics[width=0.3\textwidth]{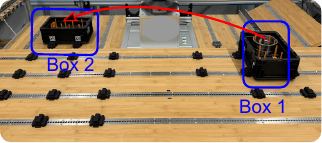}& \xSymbol& \xSymbol& \greencheckmark \\
			\cmidrule{2-5}
			& \textbf{Success rate:} & 60 \% & 60 \% & 93.33 \% \\
			\bottomrule[0.12em]	
		\end{tabular}
	\end{center}
\end{table*}
	
\end{document}